\documentclass{article}

\usepackage[preprint]{neurips_2026}

\usepackage[utf8]{inputenc} 
\usepackage[T1]{fontenc}    
\usepackage{graphicx}       
\usepackage{subcaption}     
\usepackage{hyperref}       
\usepackage{url}            
\usepackage{booktabs}       
\usepackage{algorithm}      
\usepackage{algpseudocode}  
\usepackage{amsmath}        
\usepackage{amssymb}        
\usepackage{amsfonts}       
\usepackage{nicefrac}       
\usepackage{microtype}      
\usepackage{xcolor}         
\title{Learning an Interior Layout Policy in a Domain Specific Language Action Space}

\author{%
  \textbf{Yuhao Lu \quad Weichen Zhang \quad Wenyi Xiao \quad Haohui Chen \quad Yiyun Fei} \\
  \textbf{Taobao \& Tmall Group of Alibaba}
}

\begin{document}

\maketitle

\begin{abstract}
Indoor scene layout generation is a challenging task in interior design. Existing methods often oversimplify the task by reducing room conditions to coarse 3D bounding boxes and neglecting structural elements such as doors and windows. More fundamentally, many prior approaches formulate spatial reasoning as direct coordinate prediction, thereby casting interior layout design as continuous regression over raw geometric parameters, which hinders the model from learning the underlying reasoning logic of intelligent layout design. We propose \textbf{LayoutDSL}, a novel LLM-based framework for learning an interior layout policy in a domain-specific language (DSL) action space. The DSL provides an explicit symbolic representation of layout information and serves as a structured action space for layout reasoning, where each action corresponds to an interpretable design decision. Under this DSL-based policy learning paradigm, we construct 3D-FrontDSL, a dataset of room-structure annotations paired with synthetic DSL action sequences for supervised fine-tuning. To promote a more generalizable and scalable policy with verifiable feedback, we design rewards grounded in interior design principles and physical plausibility, and optimize the policy via reinforcement learning. Extensive experiments demonstrate that LayoutDSL substantially improves spatial plausibility and design logicality over strong baselines and existing methods.

\end{abstract}    
\section{Introduction}
\label{sec:intro}

\begin{figure}[t]
\centering
\includegraphics[width=\linewidth]{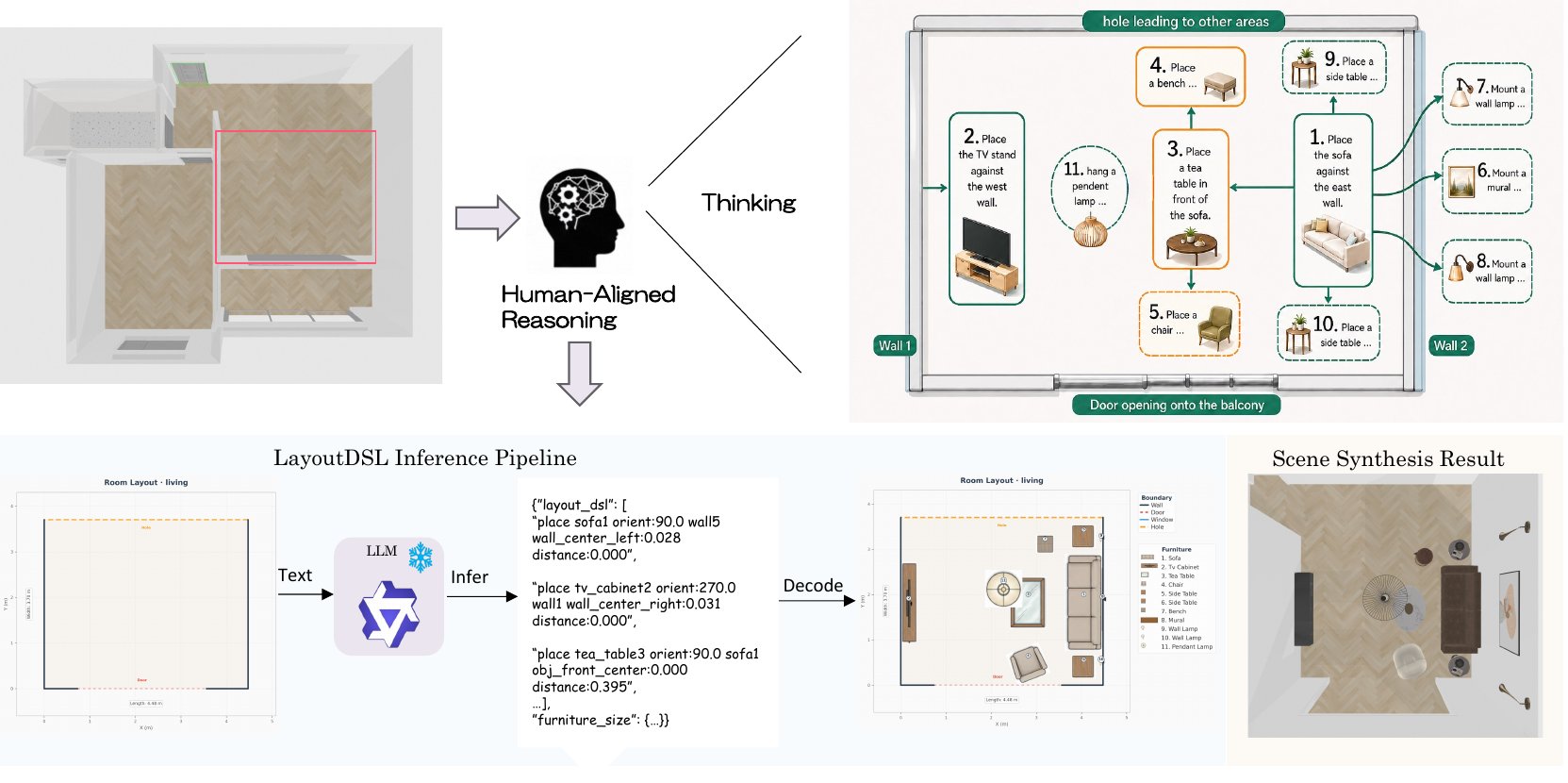}
\caption{Reasoning paradigm of \textsc{LayoutDSL}. Given a pre-partitioned meta-room with known boundaries, the LLM sequentially predicts DSL actions to mimic human-like arrangement behavior, which are then decoded into layout parameters by a DSL interpreter for downstream scene synthesis.}
\label{fig:first}
\end{figure}

Indoor scene layout generation is a central challenge in 3D indoor scene synthesis. Given a room and its structural constraints, the task is to produce a spatial arrangement that is both physically feasible and functionally meaningful. It has important applications in simulation-based data generation and automated interior design. Previous learning-based methods (~\cite{paschalidou2021atiss, tang2024diffuscene, wang2021sceneformer, yang2024physcene, wei2023lego}) typically employ diffusion-based or autoregression-based architectures to directly regress layout parameters, which exhibit limited scalability and extensibility in practical applications. Recent LLM-based works (~\cite{feng2023layoutgpt,fu2024anyhome,ccelen2024design,yang2024holodeck,aguina2024open,sun2025layoutvlm, yang2025llm, littlefair2025flairgpt, huang2025fireplace, pun2025hsm}) leverage the world knowledge of large language models to improve the generalization of layout generation. 

\begin{figure}[t]
\centering
\includegraphics[width=\linewidth]{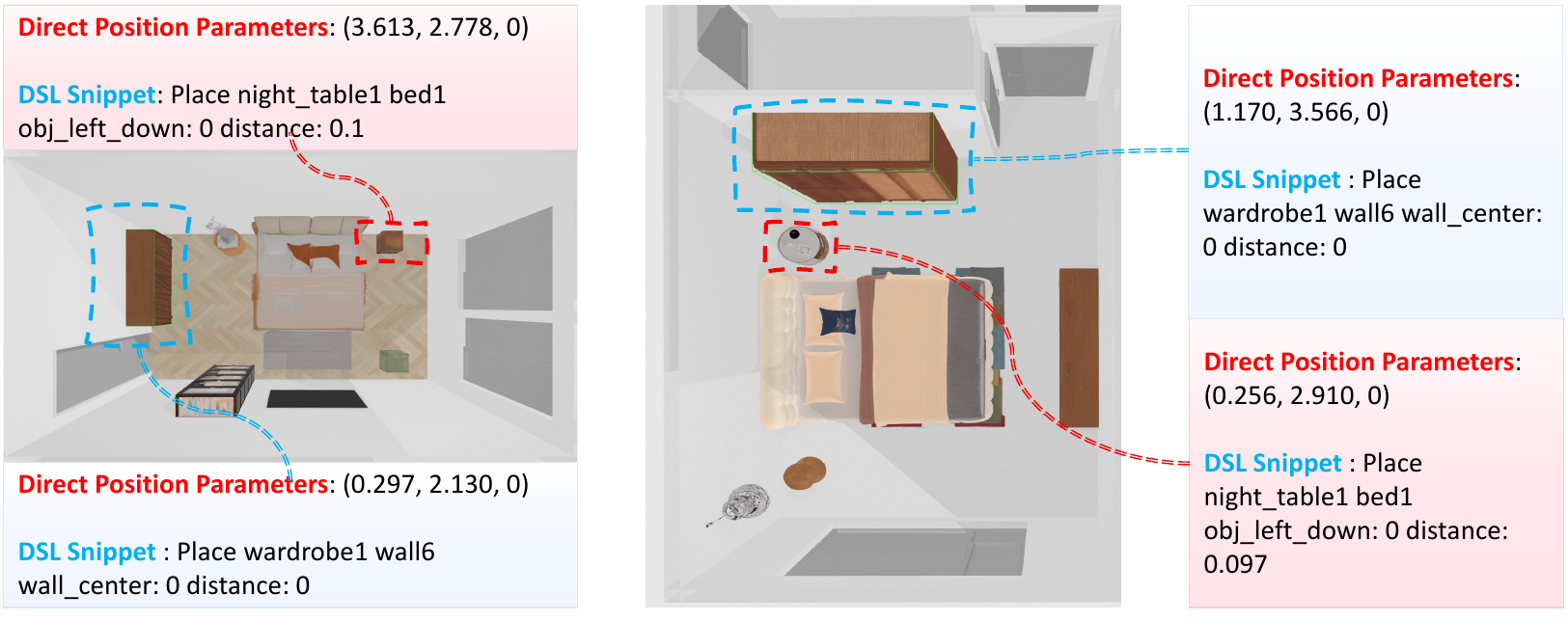}
\caption{Direct position parameters show high variance across rooms, while DSL introduces structural priors that yield more stable positional descriptions.}
\label{fig:dppdsl}
\end{figure}

However, existing methods are constrained by their modeling paradigm. Many of them reduce the room to a coarse bounding box, omitting structural elements such as doors and windows that are critical for realistic layouts. More fundamentally, they cast layout generation as continuous parameter prediction, turning a structured, constraint-driven design problem into point estimation in Euclidean space. At a comparable training data scale, direct parameterization makes it difficult for the model to develop emergent layout intelligence, as object coordinates vary substantially across rooms depending on room geometry and context, as shown in Fig.~\ref{fig:dppdsl}. In contrast, DSL incorporates structural priors over spatial relations, resulting in more stable positional descriptions.

To address the aforementioned problems, we introduce \textbf{LayoutDSL}, a novel LLM-based framework for indoor scene layout generation that projects spatial computation and layout reasoning into the action space of a domain-specific language (DSL). Figure~\ref{fig:first} outlines the modeling paradigm behind our layout reasoning pipeline. We design a layout DSL that symbolically encodes spatial relationships and geometric constraints, together with a generator–interpreter system for bidirectional conversion between layout parameters and DSL statements. And we construct \textbf{3D-FrontDSL}, a new layout dataset that incorporates key architectural elements—such as walls, doors, windows, and holes—together with corresponding layout DSL statements. Based on these paired annotations, we perform supervised fine-tuning (SFT) to align the LLM with the syntax and semantics of the layout DSL and to learn structured DSL action sequences for layout reasoning. 
To further enhance the layout quality and foster a more robust exploration mechanism, we incorporate reinforcement learning (RL) with verifiable rewards grounded in geometric feasibility and design constraints. Specifically, we construct a comprehensive reward signal by combining established geometric metrics (Collision, Out‑of‑bounds, Reachability) from prior work (~\cite{ccelen2024design, yang2024physcene, tam2025sceneeval}) with design‑aware criteria (Forbidden‑placement, Space Logicality), and update policy via Group Relative Policy Optimization (GRPO).

We conduct extensive comparative experiments to demonstrate that, after DSL-based fine-tuning and RL, a 4B-parameter LLM achieves significant layout performance gains over the baseline and surpasses much larger industry-leading LLMs that are evaluated in an unfine‑tuned, few‑shot setting. We also present comparative analyzes showing our method's advantages over existing approaches across a comprehensive set of evaluation metrics.
To summarize, our contributions are three-fold:
\begin{list}{\textbullet}{%
    \setlength{\leftmargin}{1.25em}%
    \setlength{\labelsep}{0.4em}%
    \setlength{\itemsep}{0pt}%
    \setlength{\parsep}{0pt}%
    \setlength{\topsep}{2pt}%
}
\item We design a layout domain-specific language and develop a DSL generator–interpreter system, turning indoor scene layout generation into a policy-learning problem over a structured and interpretable action space.
\item We construct \texttt{3D-FrontDSL}, a new layout dataset with room-structure annotations and Layout DSL statements for room-structure-conditioned layout generation.
\item We propose comprehensive and verifiable layout rewards that incorporate design principles, and show that reinforcement learning with these rewards substantially improves layout generation performance.
\end{list}
\section{Related Works}
\label{sec:related}

\subsection{3D Indoor Scene Layout Generation}
Improving layout generation capabilities is the central challenge in the 3D indoor scene synthesis pipeline. Traditional methods (~\cite{merrell2011interactive, weiss2018fast, infinigen2023infinite, infinigen2024indoors}) generate arrangements based on computational geometry and rule-based procedural generation. Recent popular layout generation approaches can be divided into two categories: learning-based methods (~\cite{paschalidou2021atiss, tang2024diffuscene, wang2021sceneformer, yang2024physcene, wei2023lego, sun2025hierarchically, zhai2023commonscenes, wang2018deep, ritchie2019fast}) that directly regress layout parameters using diffusion-based or autoregression-based architectures, and LLM-based methods (~\cite{yang2025sceneweaver, fu2024anyhome, ccelen2024design, aguina2024open, yang2024holodeck, hu2024scenecraft, sun2025layoutvlm, wang2024chat2layout, yang2024llplace, feng2023layoutgpt, pun2025hsm}) that leverage the inherent knowledge of LLMs to perform layout reasoning.
Learning-based approaches often have limited generalization and extensibility, whereas LLM-based approaches offer a promising alternative by leveraging the world knowledge and compositional reasoning capabilities of large language models. Some recent works (~\cite{yang2025llm, ran2025direct, bucher2025respace}) explore enhancing the layout reasoning capabilities of LLMs during post-training stages.

\subsection{Scene Layout Representation and Domain-Specific Language}
Scene layout representation plays a pivotal role in scene layout generation. The representation in existing approaches can be grouped into three paradigms: direct layout parameter, scene graph, and domain-specific language(DSL).
Direct layout parameters represent scenes as collections of numerical attributes and are widely used in learning-based models (~\cite{paschalidou2021atiss, tang2024diffuscene, wang2021sceneformer, yang2024physcene, wei2023lego, sun2025hierarchically, zhang2020deep}) and many LLM-based approaches (~\cite{ran2025direct, feng2023layoutgpt, ccelen2024design, yang2024llplace, bucher2025respace, yang2025llm}).
Scene graphs encode layouts as relational structures, where nodes denote scene objects and edges represent semantic or spatial relationships, and have been adopted to explicitly model relational dependencies in several works (~\cite{lin2024instructscene, sun2025hierarchically, yang2024holodeck, hu2024scenecraft, gao2024graphdreamer}). In contrast, DSLs formalize layouts as executable symbolic programs composed of human-interpretable primitives, and are used in works that explicitly design a DSL or grammar for layout representation (~\cite{tang2026spatialgrammar,aguina2024open, avetisyan2024scenescript, qi2018human}) as well as in action-based layout languages derived from programmatic functions (~\cite{fu2024anyhome, wang2024chat2layout}).

\subsection{Reinforcement Learning for Large
Reasoning Models}
Recent advances in reinforcement learning (RL) (~\cite{zhang2025survey, chu2025sft, guo2024deepseek, guo2025deepseek, shao2024deepseekmath, mu2024rule, peng2025lmm, ke2025survey, kumar2025llm, xu2024dpo}) have significantly improved the reasoning capabilities of large language models (LLMs), enabling multi-step problem solving through iterative feedback or reward-driven optimization. RL from Human Preferences (RLHF) (~\cite{christiano2017deep, ouyang2022training, ziegler2019fine}) and RL with Verifiable Rewards (RLVR) (~\cite{lambert2024tulu, guo2024deepseek, shao2024deepseekmath, su2025crossing}) have emerged as the dominant post-training paradigms. In indoor scene layout generation, RL methods differ in both optimization strategy and architectural design: OptiScene (~\cite{yang2025llm}) exploits implicit human preference signals to distinguish high- and low-quality layouts, DirectLayout (~\cite{ran2025direct}) proposes a CoT-grounded generative layout reward to assess layout plausibility using a VLM and a reasoning LLM, and ReSpace (~\cite{bucher2025respace}) adopts preference optimization with verifiable rewards for object addition and removal.
\section{Methodology}
\label{sec:methodology}

\subsection{Overview}
\label{ssec:overview}

\begin{figure*}[t]
\centering
\includegraphics[width=\textwidth]{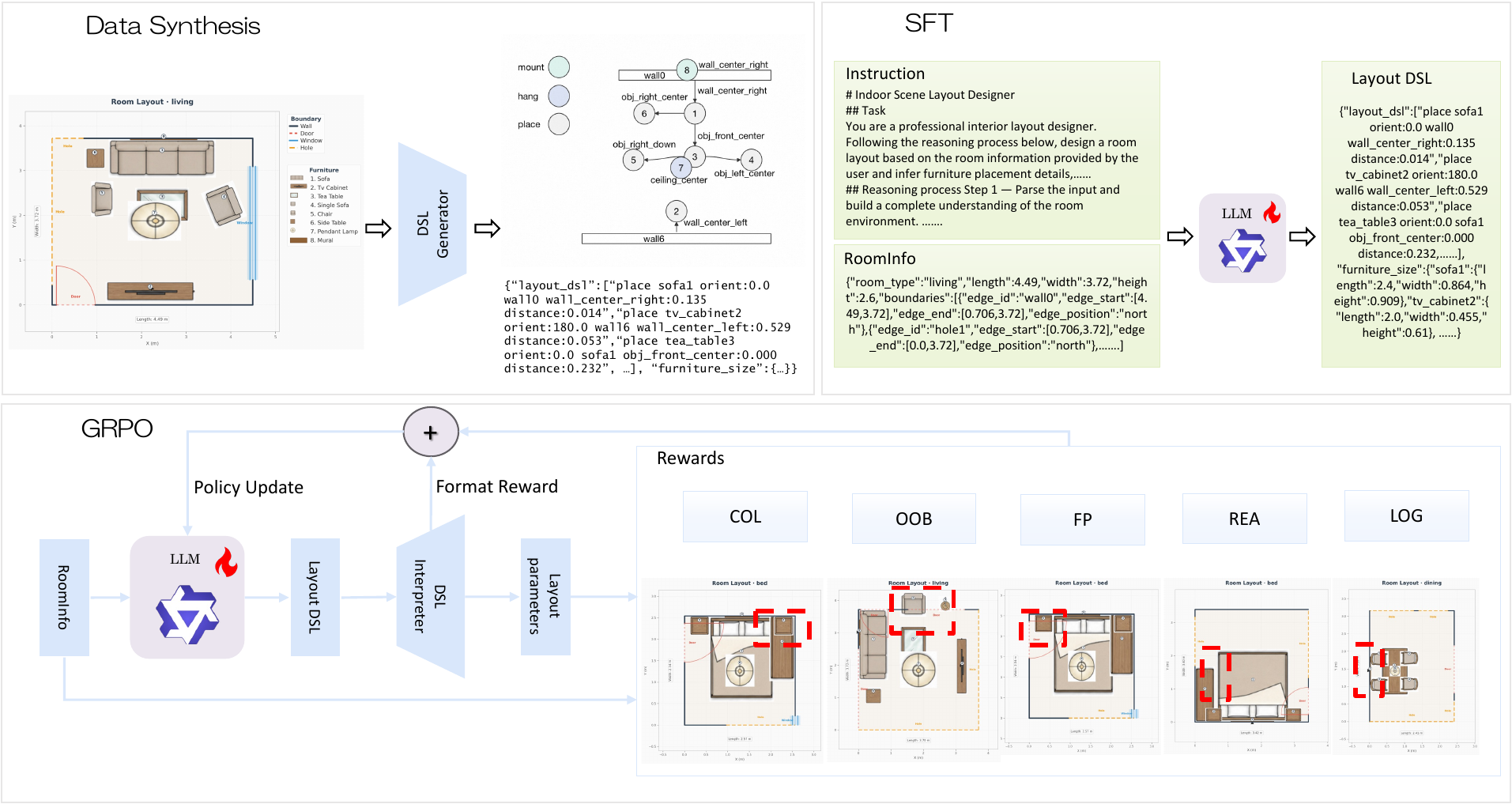}
\caption{Overview. \textbf{Data Synthesis}: Raw layout data is processed by a DSL generator that identifies spatial relationships among furniture items and converts them into structured layout DSL representations. \textbf{SFT}: The synthesized Room–Layout DSL pairs are used to fine-tune the LLM, with an example input–output pair illustrated. \textbf{GRPO}: Generated layouts are assessed across five key dimensions (the red boxes highlight regions that violate the corresponding rule); these rewards, combined with a format reward, drive GRPO-based policy updates.}
\label{fig:overview}
\end{figure*}

We present \textsc{LayoutDSL}, a novel LLM-based framework that learns an interior layout policy in a structured and interpretable DSL action space. Given meta-room structure and room openings such as doors, windows, and holes, the model learns spatial reasoning to generate physically plausible room layouts from empty rooms. An overview of \textsc{LayoutDSL} is provided in Figure~\ref{fig:overview}. The model is trained in two stages: (1) \textbf{Alignment with Layout Domain-Specific Language} in Section~\ref{ssec:dsl}, which maps geometric parameters to explicit semantic relational expressions and enables the LLM to learn layout reasoning in the DSL space; and (2) \textbf{RL with Verifiable Rewards} in Section~\ref{ssec:rl}, which optimizes the policy with rule-based rewards derived from interior design principles via Group Relative Policy Optimization (GRPO).

\subsection{Problem Formulation}
\label{ssec:problem}
We revisit the task of indoor scene layout generation, which aims to produce a complete furniture arrangement from an empty room. We define the concept of a meta-room: for any arbitrarily irregular-shaped input floor plan, each room is decomposed into one or more rectangular meta-rooms (e.g., living area, dining area) using a computational geometry algorithm.
Different from prior work that typically models rooms as coarse bounding boxes, our input preserves the true room boundary geometry and structural parameters, including doors, windows, and holes.
Formally, let a meta-room be represented as $ \mathcal{R} = (T, L, W, H, \mathcal{B}) $, where $T$ denotes the room type, $L$, $W$, and $H$ are its length, width, and height. The boundary set $\mathcal{B} = \{b_1, b_2, \dots, b_m\}$ encodes all wall segments and architectural openings along the room perimeter. Each boundary element $b_i$ is defined by:
\begin{list}{\textbullet}{%
    \setlength{\leftmargin}{1.25em}%
    \setlength{\labelsep}{0.4em}%
    \setlength{\itemsep}{0pt}%
    \setlength{\parsep}{0pt}%
    \setlength{\topsep}{2pt}%
}
\item an \texttt{edge\_id} indicating its semantic type and instance index,
\item a start point $\mathbf{p}^{\text{start}}_i \in \mathbb{R}^2$ and end point $\mathbf{p}^{\text{end}}_i \in \mathbb{R}^2$ specifying its 2D coordinates in the floor plane,
\item an \texttt{edge\_position} (e.g., \texttt{north}, \texttt{south}, \texttt{east}, \texttt{west}) indicating the orientation it belongs to.
\end{list}
Given a meta-room $\mathcal{R}$, we train an LLM to learn the conditional distribution $p_\theta(\mathcal{S} | \mathcal{R})$, where $\mathcal{S} = (\mathcal{L}_{\text{dsl}}, \mathcal{F}_{\text{s}})$ denotes the layout DSL statements and the corresponding furniture sizes. Subsequently, a DSL interpreter translates $\mathcal{S}$ into concrete layout parameters $\mathcal{L} = \{ \mathbf{l}_1, \mathbf{l}_2, \dots, \mathbf{l}_n \}$, where each $\mathbf{l}_i = ({category}_i, x_i, y_i, z_i, l_i, w_i, h_i, \theta_i)$ denotes the category, 3D position, sizes, and orientation of the $i$-th furniture item.

\subsection{Alignment with Layout Domain-Specific Language}
\label{ssec:dsl}

We devise a layout domain-specific language (DSL) to encode layout information, together with a bidirectional transformation system that enables reversible conversion between layout parameters and layout DSL statements. Concretely, each layout DSL statement is a natural-language sentence composed of symbolic tokens that explicitly specify the exact position of a piece of furniture in the current room. The system has two modules: a generator that turns 3D position coordinates and orientations into layout statements and an interpreter that performs the reverse conversion. Based on the generator, we synthesize the 3D-FrontDSL dataset, which contains raw scene layouts, room-structure annotations, and layout DSL statements. Then, we steer the LLM to achieve cognitive alignment within the semantic space defined by the layout DSL via supervised fine-tuning (SFT).

\subsubsection{Layout Language Design}
Inspired by previous works (~\cite{fu2024anyhome, wang2024chat2layout}) that employ placement functions to indirectly compute furniture positions, we aim to represent furniture positions and orientations in linguistic form comprehensively and equivalently. The core design principle of our layout DSL is as follows: given a target furniture item and the room layout, we first identify an anchor instance (either a wall or an already-placed furniture item), then classify the alignment relation between the target furniture and its anchor, and finally estimate associated alignment and distance values.

The design of the layout DSL comprises two main aspects:
(1) the top-level syntactic schema, which defines the basic structure of layout DSL statements. (2) the specification of a finite set of elements, each associated with a predefined vocabulary. Specifically, a layout DSL statement is composed of six elements:
\begin{center}
\small
\texttt{Statement = action + instance + orient:$\theta$ +
anchor + align:$\delta$ + distance:$d$}
\end{center}

where:
\begin{list}{\textbullet}{%
    \setlength{\leftmargin}{1.25em}%
    \setlength{\labelsep}{0.4em}%
    \setlength{\itemsep}{0pt}%
    \setlength{\parsep}{0pt}%
    \setlength{\topsep}{2pt}%
}
\item \texttt{action} denotes the placement manner (e.g., \texttt{place} for floor-standing furniture).
\item \texttt{instance} denotes the furniture object to be placed, specified as a concatenation of its category name and a unique numeric ID (e.g., \texttt{bed0}, \texttt{chair1}).
\item \texttt{orient} denotes the furniture's orientation, where $\theta$ represents the orientation angle in degrees.
\item \texttt{anchor} refers to an anchor instance that serves as the spatial reference for placing the furniture. Anchor instances fall into two categories: (1) \textit{wall anchors}, derived from room geometry (e.g., \texttt{ceiling0}, \texttt{wall1}), and (2) \textit{furniture anchors}, drawn from previously placed furniture instances.
\item \texttt{align} specifies the alignment type, with 26 predefined options capturing the relative spatial relationships between the furniture and its anchor; $\delta$ denotes the corresponding alignment offset (in meters). 
\item \texttt{distance} indicates the separation between the instance and its anchor; $d$ denotes the separation value (in meters).
\end{list}
Additional grammar details and the DSL generator-interpreter system are provided in the supplementary Section~\ref{sec:Syntax}.

\subsubsection{3D-FrontDSL Dataset}
\label{sssec:dataset}
To enable room-structure-conditioned layout generation and training with layout DSL, we construct the 3D-FrontDSL dataset from raw scenes in 3D-Front~\cite{fu20213d}. We first compute each room’s floor plan and extract door and window information from wall, door, and window geometries in the mesh. To alleviate the challenges posed by irregular floor plans, we adopt a hierarchical generation strategy that partitions each room into rectangular canonical placement zones; for example, a living-dining room can be split into separate living and dining zones, and an L-shaped bedroom can be transformed into a rectangular primary placement zone. We use a computational-geometry partitioning algorithm to subdivide irregular rooms, and treat any boundary shared with an adjacent region as a hole. We refer to each placement zone as a meta-room, which serves as the input unit for layout inference. We then extract the original furniture layouts corresponding to each meta-room and filter out cases with too few or too many items, as well as layouts with boundary violations. Finally, using the DSL generator described above, we generate layout DSL statements for each meta-room. In total, the resulting dataset contains \textbf{9305 meta-room layout pairs} with room-structure annotations and synthetic layout-DSL statements. Dataset statistics are provided in the supplementary Section~\ref{sec:dataset_supple}.

\subsubsection{Supervised Fine-tuning with Layout DSL}
We use supervised fine-tuning (SFT) to learn an interior layout policy in the DSL action space. The training pairs consists of $(\mathcal{R},\mathcal{S})$ from the 3D-FrontDSL, where $\mathcal{R}$ is the room structural information and $\mathcal{S}$ is the layout DSL statements with corresponding furniture sizes. During training, the LLM is prompted to generate $\mathcal{S}$ given $\mathcal{R}$ as input. The system prompt serves as a cold-start instruction, guiding the model to generate structured DSL actions that implicitly encode furniture selection, size estimation, and spatial layout reasoning. The complete prompts are provided in the supplementary Section~\ref{sec:Syntax}.

\subsection{RL with Verifiable Rewards}
\label{ssec:rl}
Although the fine-tuned model can generate coherent spatial layouts, it may still suffer from positional or circulation conflicts. We therefore leverage direct feedback from physical constraints and design principles, and use Group Relative Policy Optimization (GRPO) to further align the model with rule-based interior design preferences. Indoor scene layout generation is inherently verifiable, and previous works~\cite{tam2025sceneeval, yang2024physcene, ccelen2024design} have proposed rule-based metrics such as collision, out-of-bounds, and reachability. Building on these, we introduce additional metrics to capture room structural constraints. The resulting reward function consists of five parts: Collision (COL), Out-of-Bounds (OBB), Forbidden-Placement (FP), Route Reachability (REA), and Space Logicality (LOG), each normalized to $[0,1]$ given layout parameters $L$ and room parameters $R$.

\paragraph{Collision Reward (COL).}
The COL reward penalizes unreasonable pairwise bounding‑box overlaps between furniture items. For a layout $\mathcal{L} = \{\mathbf{l}_1, \dots, \mathbf{l}_n\}$, let $\mathcal{P} \subseteq \{(i,j) \mid i < j\}$ denote the set of furniture pairs in which both items are subject to collision detection and whose overlap ratio exceeds a context-aware threshold $\tau_{ij}$. For each $(i,j) \in \mathcal{P}$, we compute the 2D overlap ratio $o_{ij} \in [0,1]$ between the projections of their oriented bounding boxes onto the floor plan. The collision reward score is then:
\begin{equation}
    r_{\text{COL}} = (1 - \max_{(i,j)}\left( \mathbf{1}_{\{o_{ij} > \tau_{ij}\}},\, o_{ij} \right)) \cdot \min\left( \frac{|\mathcal{L}|}{N},\, 1 \right)
\end{equation}
where $|\mathcal{L}|$ denotes the number of furniture items in the layout and we set $N=7$. The weighting factor $\min\left( \frac{|\mathcal{L}|}{N},\, 1 \right)$ acts as an anti-hacking penalty to discourage the model from evading collision penalties by generating overly sparse layouts with too few objects.

\paragraph{Out-of-Bound Reward (OOB).}
The OOB reward encodes a binary criterion that penalizes layouts in which any furniture item extends beyond the room boundary. Let \(a_{\mathrm{out}}^{(k)}\) denote the outside area of the \(k\)-th furniture item and \(a^{(k)}\) its total area. The reward is 1 if all items satisfy the boundary constraint within a relative tolerance \(m\), and 0 otherwise:
\begin{equation}
r_{\mathrm{OOB}} \;=\; \prod_{k} \mathbf{1}\!\left\{ \frac{a_{\mathrm{out}}^{(k)}}{a^{(k)}} \le \epsilon_{OOB} \right\}
\end{equation}
where we set $\epsilon_{OOB}=0.01$.

\paragraph{Forbidden-Placement Reward (FP).}
The FP reward is designed specifically for room-structure-conditioned layout generation, penalizing layouts that place furniture in forbidden regions. Specifically, We enforce two semantic constraints: (i) no furniture may occupy the rectangular clearance zone in front of any door, and (ii) tall cabinets (e.g., wardrobes or bookshelves) are prohibited from the frontal zones of windows. Let \(a_{\mathrm{forb}}^{(k)}\) denote the overlap area between the \(k\)-th furniture item and forbidden regions, and \(a^{(k)}\) its total area. The reward equals 1 if all items satisfy the forbidden-placement constraints within a relative tolerance \(\epsilon\), and 0 otherwise:
\begin{equation}
r_{\mathrm{FP}} \;=\; \prod_{k} \mathbf{1}\!\left\{ \frac{a_{\mathrm{forb}}^{(k)}}{a^{(k)}} \le \epsilon_{FP} \right\}
\end{equation}
where we set \(\epsilon_{FP} = 0.01\).

\paragraph{Route Reachability Reward (REA).}
Inspired by the works (~\cite{yang2024physcene, tam2025sceneeval}), the REA reward integrates walkability and object accessibility into a unified reachability measure. First, we compute the \emph{walkable score} as a binary indicator of global connectivity: it is 1 if the unoccupied floor area forms a single connected component, and 0 otherwise. 
Second, we evaluate \emph{object accessibility score}, which measures the ratio of non-occupied area to the total area on its designated functional-side rectangle. The functional side configurations for each furniture category are automatically determined via offline LLM prompting.
The overall REA reward is defined as follow:
\begin{equation}
r_{\mathrm{REA}} = \mathbf{1}\!\left\{ \frac{A_{\text{max}}}{A_{\text{free}}} \geq \tau_{\text{walk}} \right\} \cdot \frac{1}{K} \sum_{k=1}^{K} \frac{a_{\text{free}}^{(k)}}{a_{\text{func}}^{(k)}} \cdot \min\left( \frac{|\mathcal{L}|}{N},\, 1 \right)
\end{equation}
where the first term corresponds to the \emph{walkable score} with $\tau_{\text{walk}} = 0.99$, the second term represents the \emph{object accessibility score}, and the third is an anti-hacking penalty weight (as used in COL) that prevents the model from circumventing reachability constraints by generating fewer furniture items to artificially inflate object accessibility.

\paragraph{Space Logicality Reward (LOG).}
The LOG reward evaluates layout plausibility based on interior design principles commonly followed by human designers, focusing on functional consistency and spatial harmony.  
First, \emph{functional consistency} checks whether the primary furniture category matches the room type; this yields a binary score of 1 if aligned and 0 otherwise.  
Second, \emph{spatial harmony} enforces room-specific layout conventions: for bedrooms, it verifies whether the bed’s orientation conflicts with the door location, yielding 0 if conflicts and 1 otherwise; for dining rooms, it ensures that the dining area has traversable pathways in both horizontal and vertical directions, yielding 1 if feasible and 0 otherwise. 
The overall LOG reward is defined as the product of the functional consistency and spatial harmony scores: \( r_{\mathrm{LOG}} = s_{\text{func}} \cdot s_{\text{harm}} \), where \( s_{\text{func}}, s_{\text{harm}} \in \{0,1\} \) are as defined above.

\section{Experiments}
\label{sec:experiments}

\subsection{Experiments Setup}

\begin{table*}[t]
\centering
\caption{Quantitative comparison with LLMs and existing layout generation methods on RealHome60.}
\label{tab:layout_results}
\setlength{\tabcolsep}{4pt}
\small
\resizebox{\textwidth}{!}{%
\begin{tabular}{@{}lccccccccccc@{}}
\toprule
Method & Params (B) & Fine-tuned & RT (s) & Success & COL & OOB & FP & REA & LOG & Mean \\
\midrule

\multicolumn{11}{@{}l}{\textit{Large Language Models}} \\
Qwen3-max (~\cite{yang2025qwen3})                 & 1000+   & No  & 12.48 & 95\% & 0.520 & 0.016 & 0.483 & 0.475 & 0.183 & 0.335 \\
Deepseek-v3 (~\cite{liu2024deepseek})               & 671     & No  & 20.78 & 95\% & 0.323 & 0.050 & 0.533 & 0.561 & 0.233 & 0.340 \\
GPT-5.1 (~\cite{openai2024gpt5systemcard})                & --     & No  & 50.28 & 98.33\% & 0.499 & 0.167 & 0.467 & 0.605 & 0.633 & 0.474 \\
Gemini-3.1 (~\cite{comanici2025gemini})            & --     & No  & 3714 & 98.33\% & 0.659 & 0.566 & 0.583 & 0.579 & 0.683 & 0.614 \\
\midrule

\multicolumn{11}{@{}l}{\textit{Existing Methods}} \\
LayoutGPT (~\cite{feng2023layoutgpt})        & --   & No & 6.73 & 100\% & 0.350 & 0.533 & 0.716 & 0.161 & 0.116 & 0.375 \\
AnyHome (~\cite{fu2024anyhome})      & --  & No  & 25.01 & 100\% & 0.521 & 0.933 & 0.633 & 0.324 & 0.216 & 0.525 \\
I-Design (~\cite{ccelen2024design})      & --  & No  & 74.49 & 100\% & 0.688 & \textbf{1.000} & 0.350 & 0.618 & 0.316 & 0.594 \\
\midrule

\multicolumn{11}{@{}l}{\textit{Baseline and Ours}} \\
Qwen3-4B-Instruct (~\cite{yang2025qwen3}) (Baseline)        & 4       & No  & 15.24 & 98.33\% & 0.365 & 0.050 & 0.400 & 0.621 & 0.350 & 0.357 \\
\textsc{LayoutDSL} (Ours) & 4       & Yes & 12.03 & 100\% & \textbf{0.967} & 0.867 & \textbf{0.733} & \textbf{0.958} & \textbf{0.867} & \textbf{0.878} \\
\bottomrule
\end{tabular}
}
\end{table*}

\subsubsection{Implementation Details}
To reduce computational overhead in business-to-consumer applications, we adopt Qwen3-4B-Instruct (~\cite{yang2025qwen3}) as the baseline model, trained on 8$\times$NVIDIA RTX 4090 (24GB) GPUs. All training steps are performed using LoRA-based fine-tuning (~\cite{hu2022lora}).
For SFT, we train on 9035 meta-room layout pairs from the 3D-FrontDSL dataset using LoRA with rank 32 and $\alpha = 64$. The model is optimized for 6 epochs with a learning rate of $1e{-4}$ and an effective batch size of 6.
For GRPO training, we sample 1000 meta-room's structural inputs from the 3D-FrontDSL dataset, each of which is used to generate 8 layout candidates during policy optimization. The KL divergence coefficient in GRPO is set to 0.05. We use a lighter LoRA configuration with rank 16 and $\alpha = 32$. The model is trained for 10 epochs, with a batch size of 4, gradient accumulation over 4 steps, and a learning rate of $1e{-5}$.

\subsubsection{Evaluation Metrics}
We manually curate a test set of 60 real-world residential rooms, denoted RealHome60, which mainly includes living rooms, dining rooms, and bedrooms with diverse spatial configurations. We evaluate layout quality using five objective metrics: COL, OOB, FP, REA, and LOG; their definitions and computation procedures are provided in Section~\ref{ssec:rl}. All metrics are normalized to $[0,1]$, with 1 indicating the best performance, and we report their mean as an overall score. Since LLMs may generate format-invalid DSL outputs during inference, we also report the success rate, i.e., the proportion of samples that can be successfully decoded by the DSL interpreter. All results are averaged at the meta-room level.

\begin{figure*}[t]
\centering
\includegraphics[width=\textwidth]{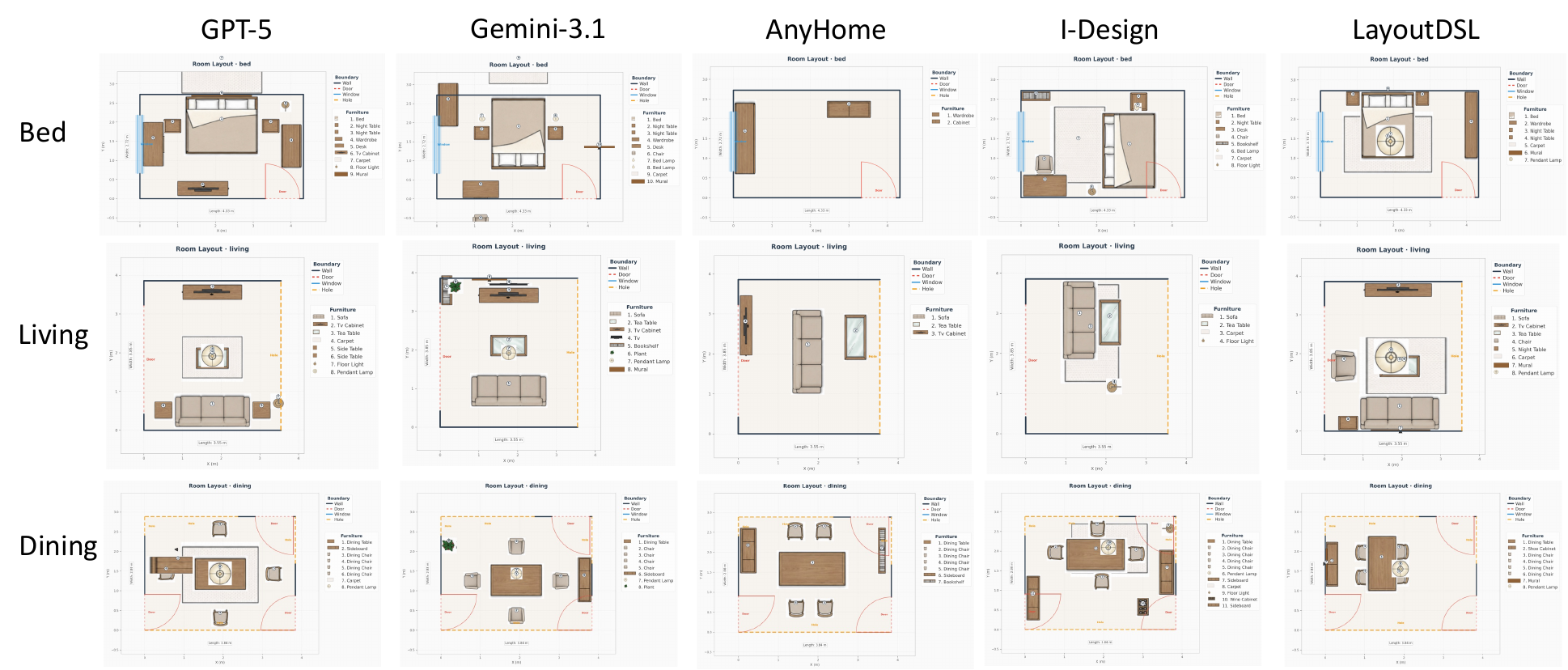}
\caption{Qualitative comparison with LLMs and existing layout generation methods in different rooms.}
\label{fig:results}
\end{figure*}

\subsection{Scene Layout Generation Results}
\paragraph{Comparison with LLMs}
We compare \textsc{LayoutDSL} with several industry-leading large LLMs~\cite{yang2025qwen3, liu2024deepseek, comanici2025gemini, openai2024gpt5systemcard}. For these models, we use the same prompts augmented with few-shot examples for layout generation. As shown in Table~\ref{tab:layout_results}, despite no fine-tuning, these LLMs achieve high success rates (95\%--98.33\%), with Gemini-3.1 obtaining the best mean score among them at 0.614. In contrast, \textsc{LayoutDSL}, trained with SFT and RL, enables a much smaller 4B model to achieve a 100\% success rate and a mean score of 0.878, outperforming Gemini-3.1 by 26.4 percentage points. Figure~\ref{fig:results} provides qualitative comparisons with GPT-5.1 and Gemini-3.1.

\paragraph{Comparison with Existing Methods}
We compare \textsc{LayoutDSL} with previous room-conditioned layout generation methods~\cite{feng2023layoutgpt, fu2024anyhome, ccelen2024design}. As shown in Table~\ref{tab:layout_results}, \textsc{LayoutDSL} achieves a mean score of 0.878, outperforming the strongest baseline, I-Design, by 28.4 points. It also surpasses all compared methods on COL, FP, REA, and LOG, while remaining competitive on OOB. The relatively weaker FP and LOG performance of prior methods is largely due to the fact that they do not explicitly model structural constraints such as doors and windows. In addition, I-Design attains a perfect OOB score because its feedback-based workflow iteratively refines layouts until out-of-bounds violations are removed, whereas \textsc{LayoutDSL} generates layouts in a fully end-to-end, single-pass manner. These results demonstrate the effectiveness of our approach for room-conditioned layout generation. Figure~\ref{fig:results} provides qualitative comparisons with AnyHome and I-Design.

\subsection{Ablation Experiments}

\begin{table}[t]
\centering
\caption{Comparison between direct parameter prediction and DSL-based generation.}
\label{tab:dpvsdsl}
\setlength{\tabcolsep}{4pt}
\small
\begin{tabular}{@{}lcccccc@{}}
\toprule
Method & COL & OOB & FP & REA & LOG & Mean \\
\midrule
Params-SFT                              & 0.633 & 0.800 & 0.649 & 0.489 & 0.650 & 0.647 \\
DSL-SFT                              & 0.715 & 0.667 & 0.716 & 0.576 & 0.750 & 0.685 \\
Params-RL                        & \textbf{0.980} & 0.850 & 0.633 & 0.899 & 0.383 & 0.749 \\
DSL-RL                        & 0.967 & \textbf{0.867} & \textbf{0.733} & \textbf{0.958} & \textbf{0.867} & \textbf{0.878} \\
\bottomrule
\end{tabular}
\end{table}

\begin{table}[t]
\centering
\caption{Ablation study on SFT, RL, and anti-hacking.}
\label{tab:ablation_sft_rl}
\setlength{\tabcolsep}{4pt}
\small
\begin{tabular}{@{}lcccccc@{}}
\toprule
Method & COL & OOB & FP & REA & LOG & Mean \\
\midrule
Qwen3-4B-Instruct & 0.365 & 0.050 & 0.400 & 0.621 & 0.350 & 0.357 \\
Qwen3-4B-SFT                 & 0.715 & 0.667 & 0.716 & 0.576 & 0.750 & 0.685 \\
\textsc{LayoutDSL} w/o anti-hacking & 0.778 & 0.933 & \textbf{0.783} & 0.737 & \textbf{0.900} & 0.825 \\
\textsc{LayoutDSL} (SFT+RL)    & \textbf{0.967} & 0.867 & 0.733 & \textbf{0.958} & 0.867 & \textbf{0.878} \\
\bottomrule
\end{tabular}
\end{table}

\paragraph{Direct Parameters vs. DSL}
We compare direct parameter prediction with DSL-based generation under the same data and training settings. As shown in Table~\ref{tab:dpvsdsl}, the DSL representation consistently outperforms direct parameters in both the SFT and RL stages. Under SFT, DSL-SFT improves the mean score from 0.647 to 0.685. The gap becomes more pronounced under RL, where DSL-RL achieves a substantially higher mean score of 0.878 compared with 0.749 for Params-RL. This indicates that the DSL action space is more effective for layout reasoning than directly predicting parameters.

\paragraph{SFT, RL, and Anti-Hacking}
We conduct an ablation study to evaluate the contributions of SFT, RL, and the anti-hacking penalty in COL and REA (Section~\ref{ssec:rl}). As shown in Table~\ref{tab:ablation_sft_rl}, the baseline model achieves a mean score of 0.357 and a low OOB score of 0.050, indicating frequent boundary violations. SFT substantially improves performance, raising the mean score to 0.685 and producing more plausible layouts. RL with rule-based rewards further boosts the mean score to 0.878, yielding the best overall physical feasibility and object placement. Removing the anti-hacking penalty degrades COL and REA, reducing the mean score to 0.825. This suggests that, without the penalty, the model tends to avoid violations by generating fewer objects, which leads to less realistic and less informative layouts.
\section{Conclusion}
\label{sec:conclusion}
In this work, we propose \textsc{LayoutDSL}, a novel paradigm for learning an interior layout policy in a domain-specific language (DSL) action space, enabling interpretable and structured layout reasoning. We further show that verifiable RL rewards can significantly improve performance.
In the future, we think that large-scale RL will facilitate layout generation models to be effectively deployed in complex intelligent design applications. Meanwhile, more challenging layout generation scenarios, such as irregular room geometries and diverse user requirements, remain important directions for future work.

{
    \small
    \bibliographystyle{ieeenat_fullname}
    \bibliography{main}
}


\appendix
\newpage
\section{Layout DSL Specification and Encoding}
\label{sec:Syntax}  

\subsection{SFT Prompt and Layout DSL Specification}
\label{ssec:sft_prompt}
We provide the complete system prompt used during supervised fine-tuning (SFT) in Table~\ref{tab:system_prompt}. This prompt explicitly defines the structured input format, denoted as \textbf{RoomInfo}, and includes a step-by-step reasoning protocol to guide the model to generate layout DSL statements.
Specifically, the prompt provides a detailed description of the syntax structure and element rules of the layout DSL, and the \textbf{Element Rules} are documented in Table~\ref{tab:dsl_syntax}, which enumerates all DSL elements along with their syntactic rules and associated vocabulary sets. 

\subsection{DSL Generator and Interpreter}
\label{ssec:system}
To achieve a reversible transformation between layout parameters and layout DSL statements, we design a reversible encoding system consisting of two components: a \textit{DSL generator}, which encodes layout parameters into layout DSL statements, and a \textit{DSL interpreter}, which decodes layout DSL statements back into layout parameters. We specify the algorithmic procedures of both the DSL generator (Algorithm~\ref{alg:dsl_generator}) and the DSL interpreter (Algorithm~\ref{alg:dsl_interpreter}), explaining how the DSL statements are encoded and decoded throughout the entire training and inference process.
The \textit{DSL generator} takes as input the room structure information and the layout parameters of the furniture items.  It first categorizes furniture items into a hierarchy of primary, secondary, and decorative types, and then sequentially generates the layout DSL for selected furniture items through a hierarchical planning process involving anchor selection, alignment detection, and attribute value computation. 
In contrast, the \textit{DSL interpreter} takes as input the layout DSL statements, the furniture dimensions, the room structure information, and outputs the numerical layout parameters. Its core mechanism analyzes the six elements of each DSL statement and computes the 3D position and orientation angle through a set of predefined element semantic mapping functions.

\begin{algorithm}[ht]
\caption{DSL Generator}
\label{alg:dsl_generator}
\begin{algorithmic}[1]
\Require Meta-room $\mathcal{R} = (T, L, W, H, \mathcal{B})$; \\
\hspace*{\algorithmicindent} Layout parameters $\mathcal{L} = \{ \mathbf{l}_1, \dots, \mathbf{l}_n \}$, with \\
\hspace*{\algorithmicindent} $\mathbf{l}_i = (\texttt{category}_i, x_i, y_i, z_i, l_i, w_i, h_i, \theta_i)$
\Ensure Layout DSL statements $\mathcal{L}_{\text{dsl}} = [d_1, \dots, d_n]$

\State $\mathcal{L}_{\text{dsl}} \gets \emptyset$, \quad $\mathcal{L}_{\text{placed}} \gets \emptyset$

\State $\mathcal{L}_{\text{seq}} \gets 
    \{ \mathbf{l}_i \in \mathcal{L} \mid \texttt{category}_i \in \text{PrimaryTypes} \} \cup
    \{ \mathbf{l}_i \in \mathcal{L} \mid \texttt{category}_i \in \text{SecondaryTypes} \} \cup
    \{ \mathbf{l}_i \in \mathcal{L} \mid \texttt{category}_i \in \text{DecorativeTypes} \}$

\For{each $\mathbf{l}_i \in \mathcal{L}_{\text{seq}}$}
    \State $\texttt{action} \gets \text{GetActionType}(\texttt{category}_i)$
    \State $\texttt{instance} \gets \texttt{category}_i +i$
    \State $\mathcal{A}_{\text{candidates}} \gets \mathcal{B} \cup \{ j \mid \mathbf{l}_j \in \mathcal{L}_{\text{placed}} \}$
    \State $\texttt{anchor} \gets \text{SelectAnchor}(\mathbf{l}_i, \mathcal{A}_{\text{candidates}})$
    \State $(\texttt{align\_type}, \delta) \gets \text{DetectAlign}(\mathbf{l}_i, \text{anchor\_id})$
    
    \State $d \gets \text{ComputeDist}(\texttt{align\_type}, \mathbf{l}_i, \texttt{anchor})$
    
    \State $d_i \gets \texttt{action} \texttt{ }\texttt{instance} \texttt{ orient:}\theta_i\texttt{ } $
    \Statex \quad $\texttt{anchor} \texttt{ } \texttt{align\_type}\texttt{:}\delta \texttt{ distance:}d$
    
    \State $\mathcal{L}_{\text{dsl}}.\text{append}(d_i)$
    \State $\mathcal{L}_{\text{placed}}.\text{add}(\mathbf{l}_i)$
\EndFor

\State \Return $\mathcal{L}_{\text{dsl}}$
\end{algorithmic}
\end{algorithm}

\begin{algorithm}[t]
\caption{DSL Interpreter}
\label{alg:dsl_interpreter}
\begin{algorithmic}[1]
\Require Layout DSL statements $\mathcal{L}_{\text{dsl}} = [d_1, \dots, d_n]$; \\
\hspace*{\algorithmicindent} Furniture sizes $\mathcal{F}_s: \texttt{id} \mapsto (l_i, w_i, h_i)$; \\
\hspace*{\algorithmicindent} Meta-room $\mathcal{R} = (T, L, W, H, \mathcal{B})$
\Ensure Layout parameters $\mathcal{L} = \{ \mathbf{l}_1, \dots, \mathbf{l}_n \}$, with \\
\hspace*{\algorithmicindent} $\mathbf{l}_i = (\texttt{category}_i, x_i, y_i, z_i, l_i, w_i, h_i, \theta_i)$

\State $\mathcal{L} \gets \emptyset$, \quad $\mathcal{F}_{\text{geom}} \gets \emptyset$

\For{each $d_i \in \mathcal{L}_{\text{dsl}}$}
    \State $(\texttt{action}, \texttt{instance}, \theta_i, \texttt{anchor},$
    \Statex \quad $[\texttt{align\_type}, \delta], d) \gets \text{Parse}(d_i)$

    \If{$\texttt{anchor} \in \mathcal{B}$}
        \State $\texttt{geom\_ref} \gets \text{Boundary}(\mathcal{B}, \texttt{anchor})$
    \Else
        \State $\texttt{geom\_ref} \gets \text{Furniture}(\mathcal{F}_{\text{geom}}, \texttt{anchor})$
    \EndIf
    \State $\text{func\_call} \gets \text{GetAlignFunc}(\texttt{align\_type})$
    \State $(x_i, y_i, z_i) \gets \text{func\_call}(\texttt{action}, \texttt{instance}, $
    \Statex $\texttt{geom\_ref}, \theta_i, \delta, d)$

    \State $\texttt{category}_i \gets \text{GetCategory}(\texttt{instance})$
    \State $\mathbf{l}_i \gets (\texttt{category}_i, x_i, y_i, z_i, l_i,w_i, h_i, \theta_i)$
    
    \State $\mathcal{L}.\text{append}(\mathbf{l}_i)$
    \State $\mathcal{F}_{\text{geom}}.\text{update}(\mathbf{l}_i)$
\EndFor

\State \Return $\mathcal{L}$
\end{algorithmic}
\end{algorithm}

\section{Dataset Statistics for the 3D-FrontDSL}
\label{sec:dataset_supple}
We provide key statistics of the 3D-FrontDSL dataset, which contains 9,305 meta-room layout pairs with room structure annotations and corresponding Layout DSL statements.

\paragraph{Room Types.}
In the 3D-FrontDSL dataset, meta-rooms are categorized into five main room types. The distribution of room type counts is as follows:
\texttt{Bedroom} (4,555), \texttt{Dining} (2,106), \texttt{Living} (1,918), \texttt{Library} (550), and \texttt{KidsRoom} (176).

\paragraph{Furniture Categories.}
We consolidate all furniture categories from the training data into a unified set of 29 categories. The set includes \texttt{bed}, \texttt{bench}, \texttt{bookshelf}, \texttt{cabinet}, \texttt{carpet}, \texttt{chair}, \texttt{children bed}, \texttt{children table}, \texttt{cloth cabinet}, \texttt{desk}, \texttt{dining chair}, \texttt{dining table}, \texttt{dresser}, \texttt{floor light}, and \texttt{mural}. It further contains \texttt{night table}, \texttt{pendant lamp}, \texttt{shoe cabinet}, \texttt{side table}, \texttt{sideboard}, \texttt{single bed}, \texttt{single sofa}, \texttt{sofa}, \texttt{tea table}, \texttt{tv cabinet}, \texttt{wall lamp}, \texttt{bed lamp}, \texttt{wine cabinet}, and \texttt{wardrobe}.

\begin{figure}[ht]
\centering
\begin{subfigure}[b]{0.48\linewidth}
  \centering
  \includegraphics[width=\linewidth]{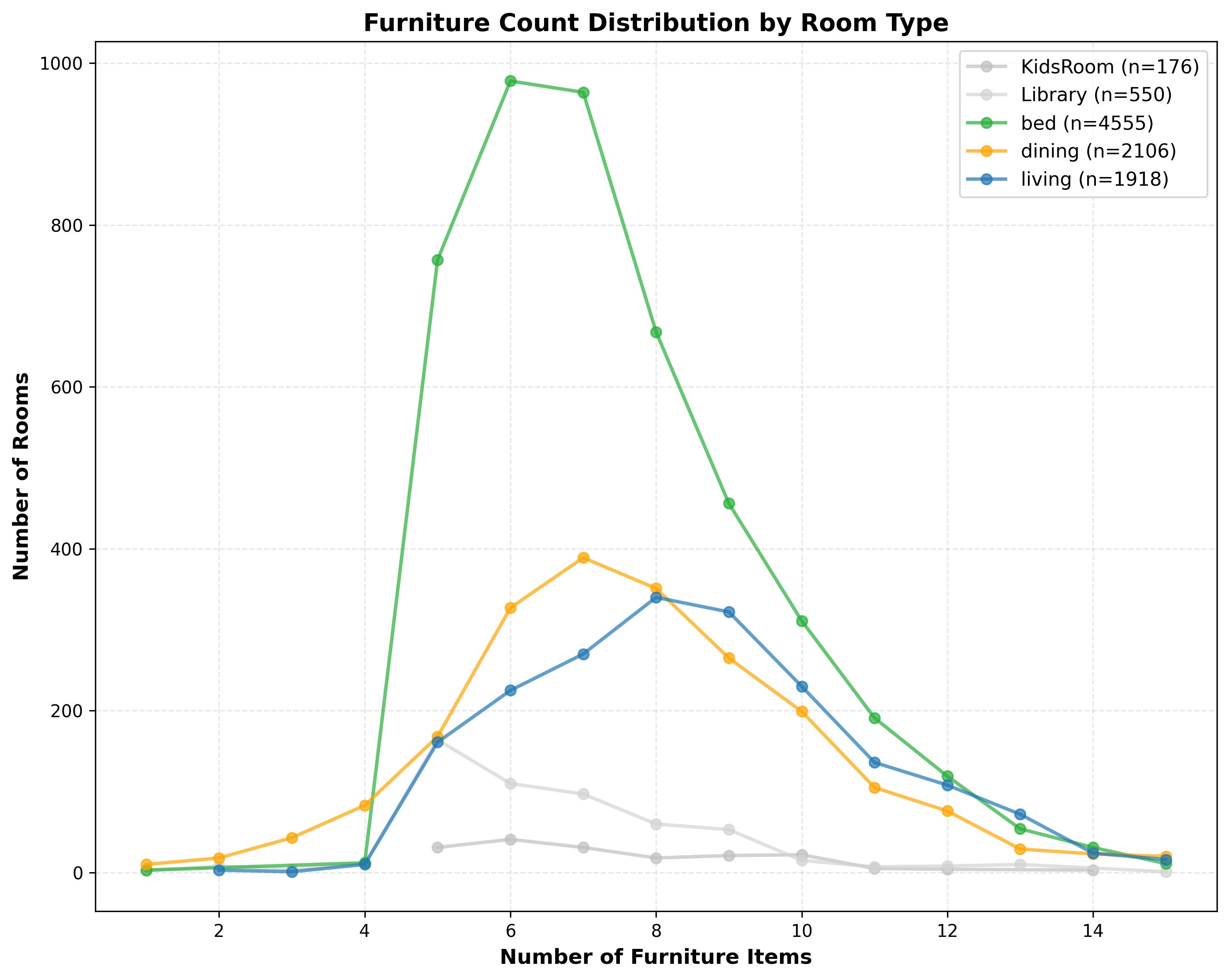}
  \caption{Furniture count distribution across different room types.}
  \label{fig:furniture_distribution}
\end{subfigure}
\hfill
\begin{subfigure}[b]{0.48\linewidth}
  \centering
  \includegraphics[width=\linewidth]{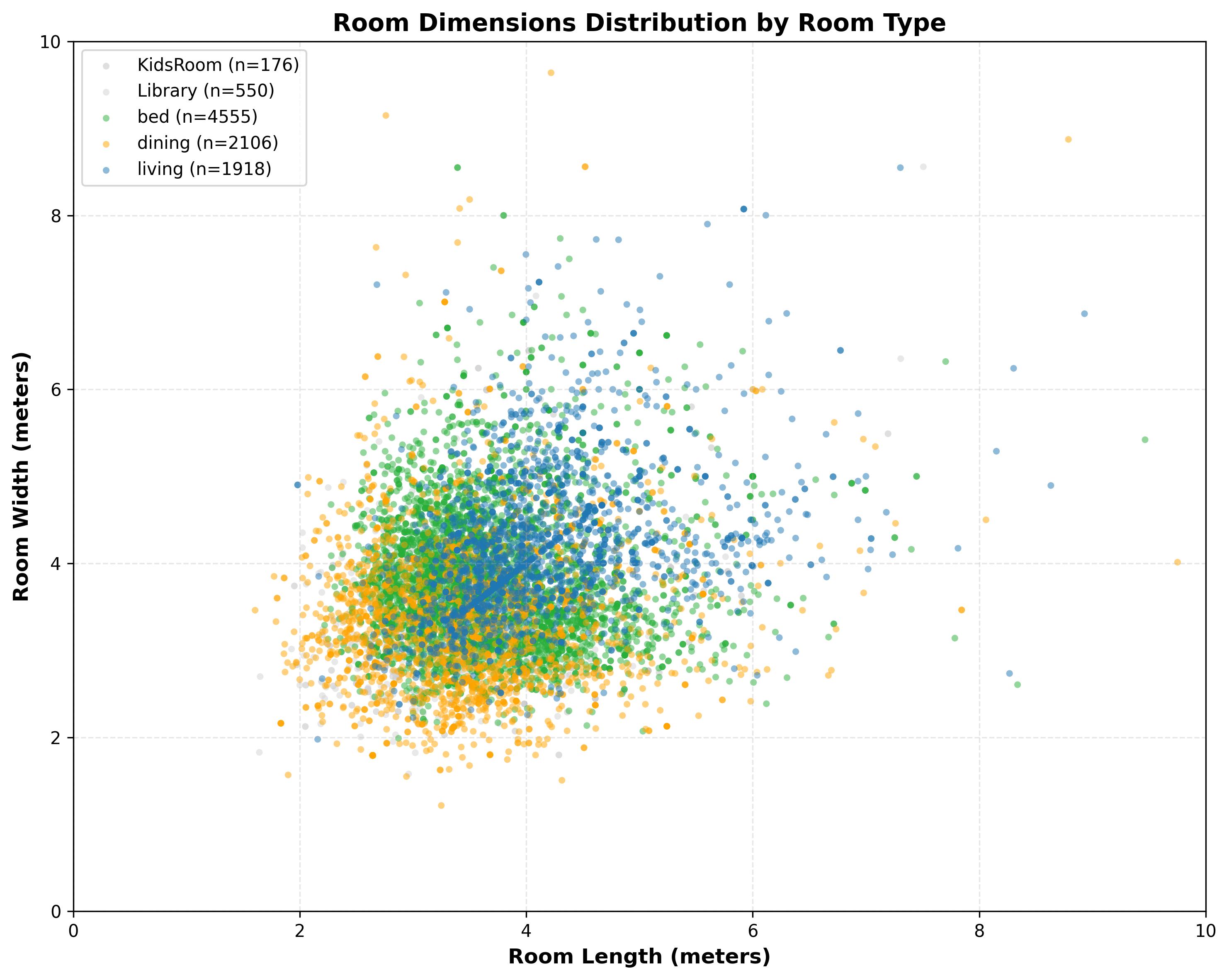}
  \caption{Room dimensions distribution across different room types.}
  \label{fig:room_dimensions_distribution}
\end{subfigure}
\caption{Dataset statistics: (a) furniture count distribution and (b) room dimensions distribution.}
\label{fig:dataset_statistics}
\end{figure}

\paragraph{Furniture Count Distribution.}
We analyze the furniture count distribution across meta-room samples to assess layout richness. As shown in Figure~\ref{fig:furniture_distribution}, the majority of meta-rooms contain between 5 and 12 furniture items. The average number of furniture items per meta-room is 7.67.

\paragraph{Room Dimensions.}  
We analyze the distribution of room length and width to assess spatial diversity in terms of floor area. As shown in Figure~\ref{fig:room_dimensions_distribution}, both length and width are primarily concentrated in the range of 2 to 6 meters, reflecting typical residential room scales. The average room length is 3.722 meters and the average width is 3.775 meters.

\paragraph{Boundary Features.}  
We examine the occurrence frequency and average segment length of boundary elements—doors, windows, holes—in meta-room annotations. These features capture the geometric and functional complexity of room layouts. On average, each room contains 0.95 doors with a length of 1.11 meters, 0.57 windows with a length of 1.18 meters, and 1.07 holes with a length of 2.55 meters. These boundary elements vary across individual rooms and significantly influence furniture placement decisions and layout feasibility.

\section{Training Details and Additional Experiments}
\subsection{Performance Enhancement via GRPO}
\label{sec: GRPO}
\begin{figure}[t]
\centering
\includegraphics[width=\linewidth]{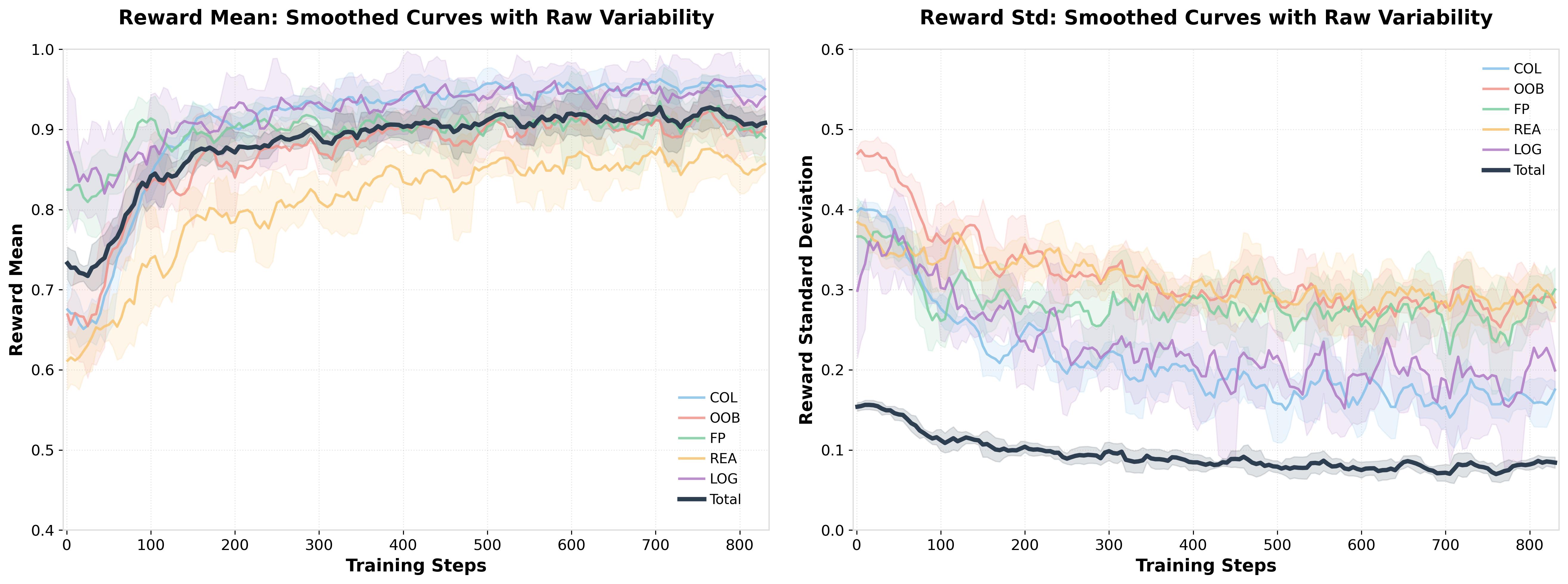}
\caption{GRPO reward curves. The left panel shows the mean values of the rewards, and the right panel depicts their variances. Total denotes the composite reward, defined as the weighted sum of the five individual rewards plus an additional format validity reward.}
\label{fig:reward}
\end{figure}

We employ Group Relative Policy Optimization (GRPO) (~\cite{guo2025deepseek}) further fine‑tune the model initialized by supervised fine‑tuning (SFT). We use an equally weighted average of the five rewards (COL, OOB, FP, REA, LOG) as the unified training signal. Additionally, the model outputs DSL statements, which are decoded into layout parameters by a DSL interpreter; if decoding fails (e.g. due to syntactic errors), the sample is assigned a total reward of 0, serving as an implicit format validity penalty.
We adopt the standard GRPO objective function, and incorporate a KL divergence penalty between the current policy and the SFT-initialized reference policy. Figure~\ref{fig:reward} shows the reward curves during GRPO training. 

\subsection{Reproduction Details for Existing Methods}
To ensure a fair comparison, we note that the previous methods do not natively support the room-boundary input setting used in our task. Therefore, when reproducing these baselines, we simplify the room information as much as possible and adapt it to the input format required by each method. We then evaluate each baseline under its original inference pipeline whenever available.
For LayoutGPT~\cite{feng2023layoutgpt}, we adapt its 3D scene synthesis pipeline to each meta-room and generate layouts using the corresponding few-shot exemplars provided by the method. The generated object categories are then mapped to our evaluation setting.
For AnyHome~\cite{fu2024anyhome}, we use its room layout and object placement generation module to produce layout parameters. Since AnyHome accepts door information as input, we additionally provide the corresponding door data in the floor plan input.
For I-Design~\cite{ccelen2024design}, we construct a room-type-specific textual prompt, specify the room dimensions, and set the number of furniture items, an input parameter of I-Design, to an integer between 5 and 12. We then obtain the final layout by executing the full multi-agent workflow of I-Design.
In all cases, we map the predicted object categories to our label space before computing the evaluation metrics.

\begin{table}[t]
\centering
\caption{Quantitative comparison on Model Scale.}
\label{tab:model scale}
\setlength{\tabcolsep}{4pt}
\small
\begin{tabular}{@{}lcccccc@{}}
\toprule
Method & COL & OOB & FP & REA & LOG & Mean \\
\midrule
Qwen3-0.6B-SFT    & 0.552 & 0.499 & 0.633 & 0.488 & 0.750 & 0.584 \\
Qwen3-1.7B-SFT    & 0.603 & 0.499 & 0.566 & 0.544 & 0.700 & 0.582 \\
Qwen3-4B-SFT    & \textbf{0.715} & \textbf{0.667} & \textbf{0.716} & 0.576 & \textbf{0.750} & \textbf{0.685} \\
Qwen3-8B-SFT   & 0.627 & \textbf{0.667} & 0.699 & \textbf{0.625} & 0.733 & 0.671 \\
\bottomrule
\end{tabular}
\end{table}

\subsection{Effect of Model Scale}
We compare SFT performance across Qwen3 models with 0.6B, 1.7B, 4B, and 8B parameters. As shown in Table~\ref{tab:model scale}, the improvement is more substantial from 0.6B to 4B, whereas the gains from 4B to 8B are relatively modest.



\section{Downstream Tasks}

\subsection{3D Indoor Scene Synthesis}
Beyond layout generation, we demonstrate \textsc{LayoutDSL}'s application in commercial scene synthesis. Our complete scene synthesis pipeline takes floor plan data as input, first partitions it into meta-rooms, and then employs \textsc{LayoutDSL} to generate layouts within each meta-room. The resulting layout information—including category, position, size, and orientation—is used to drive a lightweight model retrieval system that fetches 3D assets from a model library based on category and size. The final synthesized scenes are visualized by using a Three.js-based web frontend interface, as shown in Figure~\ref{fig:first}.

\subsection{User-Conditioned Layout Generation}
\label{sec:user}
\begin{figure*}[!t]
\centering
\includegraphics[width=0.9\textwidth]{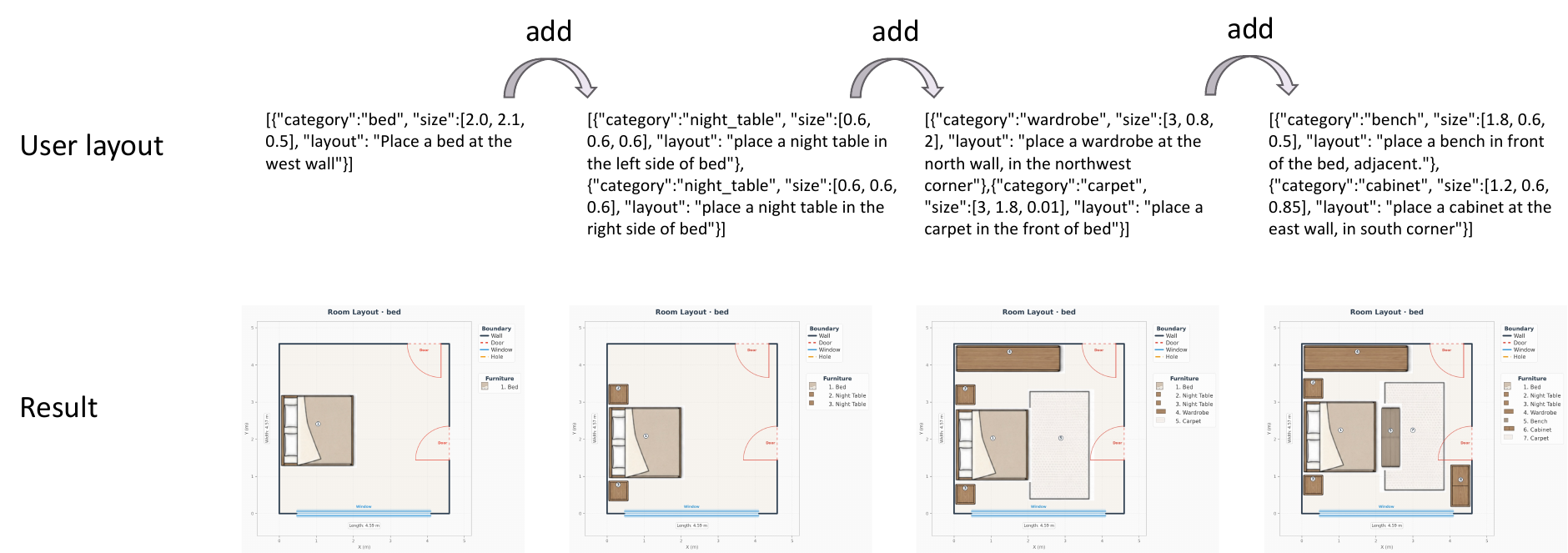}
\caption{Qualitative results for user-conditioned layout generation. Add means that the next layout accumulates new layout instructions on top of the previous layout instructions.}
\label{fig:user}
\end{figure*}

We now discuss layout generation conditioned on meta-room and user-specified layout instructions. In this setting, the model generates room layouts according to explicit natural-language instructions provided by the user. Thanks to the semantic structure of the Layout DSL, placement instructions such as ``place the sofa at the center of the west wall'' can be naturally represented and executed within our framework. 
The input to the model consists of two parts, \texttt{roominfo} and \texttt{user\_input}. The \texttt{user\_input} specifies the desired furniture items, their dimensions, and coarse placement instructions. We adapt \textsc{LayoutDSL} to this task by introducing a dedicated system prompt and several few-shot examples, without additional fine-tuning. The complete prompt template and instruction format are provided in Table~\ref{tab:system_prompt_user} and the supplementary material. 
Figure~\ref{fig:user} presents qualitative examples showing how the same room can be progressively updated according to iterative user instructions. Even with coarse instructions, the model is able to infer precise positions, orientations, and relative spatial relationships, demonstrating strong controllability in downstream layout generation.

\begin{table*}[t]
\centering
\caption{System prompt for SFT.}
\label{tab:system_prompt}
\scriptsize
\renewcommand{\arraystretch}{1.0}
\begin{tabular}{@{}p{0.15\linewidth}|p{0.8\linewidth}@{}}
\toprule
\textbf{System Prompt} &
\textbf{\# Indoor Scene Layout Designer} \\
& \textbf{\#\# Task} \\
& You are a professional interior layout designer. Following the reasoning process below, design a room layout based on the \textbf{RoomInfo} provided by the user and infer furniture placement details, including two parts: \\
& \quad 1. \texttt{layout\_dsl}: layout DSL statements that describe furniture placement. \\
& \quad 2. \texttt{furniture\_size}: furniture dimensions (length, width, height). \\
& \textbf{\#\# Reasoning process} \\
& \textbf{Step 1 --- Parse the input and build a complete understanding of the room environment} \\
& \quad Room attributes: \\
& \quad\quad -- Identify the room type and the room's length, width and height. \\
& \quad Wall information (boundaries): \\
& \quad\quad -- Identify room boundary edges: \texttt{edge\_id} is the boundary type plus an id; \texttt{edge\_position} indicates the side position; \texttt{start} and \texttt{end} are the endpoints of the boundary segment. \\
& \quad Boundary analysis: \\
& \quad\quad -- Distinguish boundary types clearly: wall, door, window, hole. This is the basis for all layout decisions. \\
& \quad Coordinate system: \\
& \quad\quad -- The room bottom-left corner is the origin (0,0); units are meters. \\
& \quad Orientation system: \\
& \quad\quad -- 0°: south \quad 90°: west \quad 180°: north \quad 270°: east \\
& \textbf{Step 2 --- Select furniture and estimate sizes} \\
& \quad Select furniture: based on room type and room size, determine a complete list of furniture to place. \\
& \quad Define sizes: assign realistic, proportionate dimensions (length, width, height) to each item in the list according to the room size. \\
& \quad Furniture size and orientation conventions: \\
& \quad\quad -- length: the side length perpendicular to the furniture's facing/orientation direction. \\
& \quad\quad -- width: the side length along the furniture's facing/orientation direction. \\
& \quad\quad -- height: vertical height. \\
& \textbf{Step 3 --- Layout reasoning} \\
& \quad Layout order: \\
& \quad\quad a. Primary furniture: first place core large furniture that directly depends on walls. \\
& \quad\quad b. Dependent furniture: next place auxiliary furniture that depends on primary furniture. \\
& \quad\quad c. Supplementary furniture: finally place functional or decorative items. \\
& \quad DSL instruction generation: \\
& \quad\quad -- Convert the layout reasoning into DSL statements, producing one DSL statement per furniture placement. \\
& \quad\quad -- DSL syntax structure: \\
& \quad\quad \texttt{action + instance + orient:theta + anchor + align:delta + distance:d} \\
& \quad\quad -- DSL syntax element rules: All generated DSL statements must strictly follow the \textbf{Element Rules}. \\
& \textbf{Step 4 --- Final output format} \\
& \quad Strictly follow the JSON structure below: \\
& \quad\quad \texttt{\{} \\
& \quad\quad\quad \texttt{"layout\_dsl": [} \\
& \quad\quad\quad\quad \texttt{"string", \quad \# example DSL statement} \\
& \quad\quad\quad\quad \texttt{...} \\
& \quad\quad\quad \texttt{],} \\
& \quad\quad\quad \texttt{"furniture\_size": \{} \\
& \quad\quad\quad\quad \texttt{"instance\_id": \{"length": float, "width": float, "height": float\},} \\
& \quad\quad\quad\quad \texttt{...} \\
& \quad\quad\quad \texttt{\}} \\
& \quad\quad \texttt{\}} \\
\bottomrule
\end{tabular}
\end{table*}

\begin{table*}[tp]
\centering
\caption{Element Rules of Layout DSL syntax.}
\label{tab:dsl_syntax}
\small
\renewcommand{\arraystretch}{1.0}
\begin{tabular}{@{}p{0.15\linewidth}|p{0.8\linewidth}@{}}
\toprule
\textbf{Element} & \textbf{Rules} \\

\midrule
\textbf{a. action} & Choose from [place, mount, hang] \\
& \quad -- place: furniture placed on the floor. \\
& \quad -- mount: attached/mounted to a wall (e.g., mural). \\
& \quad -- hang: suspended from the ceiling (e.g., pendant\_lamp). \\[0.2em]

\midrule

\textbf{b. instance} & Unique furniture ID \\
& \quad -- Format: category+number (e.g., bed0, night\_table1). \\[0.2em]
\midrule

\textbf{c. orient:theta} & theta is the angular value representing the instance's orientation. \\[0.2em]
\midrule

\textbf{d. anchor} & Must be either an edge\_id defined in the room boundaries (e.g., wall3, window1), a previously placed furniture instance (e.g., bed0), or a ceiling center (e.g., ceiling0). \\[0.2em]
\midrule

\textbf{e. align:delta} & align must be chosen from the full list of align\_type (see below). delta is a float representing an additional offset applied on top of the chosen alignment; the offset direction is determined by align\_type. \\[0.2em]

& Full list of align\_type and their meanings \\[0.2em]
& \textit{Wall-based alignment:} \\
& \quad --  wall\_center: align to the center point of the wall segment. \\
& \quad -- wall\_left\_corner: align to the wall's left endpoint. \\
& \quad -- wall\_right\_corner: align to the wall's right endpoint. \\
& \quad -- wall\_center\_left: offset to the left from the wall center by delta. \\
& \quad -- wall\_center\_right: offset to the right from the wall center by delta. \\[0.2em]
& \textit{Ceiling-based alignment:} \\
& \quad -- ceiling\_center: relative to the ceiling center; x-axis offset is d, y-axis offset is delta. \\[0.2em]
& \textit{Object left-side alignment:} \\
& \quad -- obj\_left\_down: left side of the anchor, bottom aligned. \\
& \quad -- obj\_left\_top: left side of the anchor, top aligned. \\
& \quad -- obj\_left\_center: align to the center of the anchor's left side. \\
& \quad -- obj\_left\_center\_down: offset downward from the anchor's left-side center by delta. \\
& \quad -- obj\_left\_center\_top: offset upward from the anchor's left-side center by delta. \\[0.2em]
& \textit{Object right-side alignment:} \\
& \quad -- obj\_right\_down: right side of the anchor, bottom aligned. \\
& \quad -- obj\_right\_top: right side of the anchor, top aligned. \\
& \quad -- obj\_right\_center: align to the center of the anchor's right side. \\
& \quad -- obj\_right\_center\_down: offset downward from the anchor's right-side center by delta. \\
& \quad -- obj\_right\_center\_top: offset upward from the anchor's right-side center by delta. \\[0.2em]
& \textit{Object front-side alignment:} \\
& \quad -- obj\_front\_left: in front of the anchor, left edge aligned. \\
& \quad -- obj\_front\_right: in front of the anchor, right edge aligned. \\
& \quad -- obj\_front\_center: align to the center of the anchor's front side. \\
& \quad -- obj\_front\_center\_left: offset left from the anchor's front-side center by delta. \\
& \quad -- obj\_front\_center\_right: offset right from the anchor's front-side center by delta. \\[0.2em]
& \textit{Object back-side alignment:} \\
& \quad -- obj\_back\_left: behind the anchor, left edge aligned. \\
& \quad -- obj\_back\_right: behind the anchor, right edge aligned. \\
& \quad -- obj\_back\_center: align to the center of the anchor's back side. \\
& \quad -- obj\_back\_center\_left: offset left from the anchor's back-side center by delta. \\
& \quad -- obj\_back\_center\_right: offset right from the anchor's back-side center by delta. \\

\midrule

\textbf{f. distance:d} & d is a float. The direction of the distance is determined by action and align\_type. For \texttt{mount} and \texttt{hang}, d corresponds to the z-axis distance. For \texttt{place}, d corresponds to either the x-axis or y-axis distance depending on align\_type. \\

\bottomrule
\end{tabular}
\end{table*}

\begin{table*}[t]
\centering
\caption{System prompt for user-conditioned layout generation.}
\label{tab:system_prompt_user}
\scriptsize
\renewcommand{\arraystretch}{1.0}
\begin{tabular}{@{}p{0.15\linewidth}|p{0.8\linewidth}@{}}
\toprule
\textbf{System Prompt} &
\textbf{\# Indoor Scene Layout Designer} \\
& \textbf{\#\# Task} \\
& You are a professional interior layout designer. Following the reasoning process below, design a room layout based on the room information [\texttt{\${roominfo}}] and a list of user-specified furniture items with their categories, dimensions, and natural-language placement descriptions [\texttt{\${user\_layout}}], finally infer furniture placement details, including two parts: \\
& \quad 1. \texttt{layout\_dsl}: layout DSL statements that describe furniture placement. \\
& \quad 2. \texttt{furniture\_size}: furniture dimensions (length, width, height). \\
& \textbf{\#\# Reasoning Process} \\
& \textbf{Step 1 — Understand the room information} \\
& \quad \texttt{roominfo}: \\
& \quad\quad -- Room attributes: \\
& \quad\quad\quad Identify the room type and the room's length, width and height. \\
& \quad\quad -- Wall information (boundaries): \\
& \quad\quad\quad Identify room boundary edges: \texttt{edge\_id} is the boundary type plus an id; \texttt{edge\_position} indicates the side position; \texttt{start} and \texttt{end} are the endpoints of the boundary segment. \\
& \quad\quad -- Boundary analysis: \\
& \quad\quad\quad Distinguish boundary types clearly: wall, door, window and hole. This is the basis for all layout decisions. \\
& \quad\quad -- Coordinate system: \\
& \quad\quad\quad The room bottom-left corner is the origin (0,0); units are meters. \\
& \quad\quad -- Orientation system: \\
& \quad\quad\quad 0°: south \quad 90°: west \quad 180°: north \quad 270°: east \\
& \\
& \quad \texttt{user\_layout}: \\
& \quad\quad -- User-specified furniture placement list that records the furniture the user intends to place in the room. \\
& \quad\quad -- For each furniture item, record: \\
& \quad\quad\quad \texttt{category}: furniture type/name (e.g., \texttt{'bed'}, \texttt{'night\_table'}, \texttt{'mural'}). \\
& \quad\quad\quad \texttt{size}: furniture dimensions [length, width, height] in meters. \\
& \quad\quad\quad \texttt{layout}: natural language description of the desired placement location and orientation relative to room boundaries and other furniture (e.g., \texttt{'the bed should be placed against the west wall, slightly right of center'}). \\
& \\
& \textbf{Step 2 — Interpret layout intent and resolve placement} \\
& \quad For each furniture item: \\
& \quad\quad -- Analyze the provided layout description to infer spatial relationships. \\
& \quad\quad -- Convert the layout reasoning into DSL statements, producing one DSL statement per furniture placement. \\
& \quad\quad -- DSL syntax structure: \\
& \quad\quad\quad \texttt{action + instance + orient:theta + anchor + align:delta + distance:d} \\
& \quad\quad -- DSL syntax element rules: All generated DSL statements must strictly follow the \textbf{Element Rules}. \\
& \\
& \textbf{Step 3 — Final output format} \\
& \quad Strictly follow the JSON structure below: \\
& \quad\quad \texttt{\{} \\
& \quad\quad\quad \texttt{"layout\_dsl": [} \\
& \quad\quad\quad\quad \texttt{"string", \quad \# example DSL statement} \\
& \quad\quad\quad\quad \texttt{...} \\
& \quad\quad\quad \texttt{],} \\
& \quad\quad\quad \texttt{"furniture\_size": \{} \\
& \quad\quad\quad\quad \texttt{"instance\_id": \{"length": float, "width": float, "height": float\},} \\
& \quad\quad\quad\quad \texttt{...} \\
& \quad\quad\quad \texttt{\}} \\
& \quad\quad \texttt{\}} \\
\bottomrule
\end{tabular}
\end{table*}



\end{document}